\documentclass{article}
\usepackage{icml2001}
\usepackage{epsf}
\usepackage{mlapa}
\usepackage{amssymb}
\usepackage{amsmath}
\usepackage{amsthm}

\begin{document} 

\twocolumn[

\icmltitle{A Multi-Agent, Policy-Gradient approach to Network Routing}
 
\icmlauthor{Nigel Tao}{Nigel.Tao@cs.anu.edu.au}
\icmladdress{Department of Computer Science, The Australian National
  University, Canberra 0200, Australia}

\icmlauthor{Jonathan Baxter}{jbaxter@whizbang.com}
\icmladdress{WhizBang! Labs. 4616 Henry Street Pittsburgh, PA 15213}

\icmlauthor{Lex Weaver}{Lex.Weaver@cs.anu.edu.au}
\icmladdress{Department of Computer Science, The Australian National
  University, Canberra 0200, Australia}

\vskip 0.1in
]

\newcommand{\node}{\textsf}
\newcommand{\mnode}{\mathsf}

\begin{abstract}
  Network routing is a distributed decision problem which naturally
  admits numerical performance measures, such as the average time for
  a packet to travel from source to destination. \textsc{Olpomdp}, a
  policy-gradient reinforcement learning algorithm, was successfully
  applied to simulated network routing under a number of network
  models.  Multiple distributed agents (routers) learned co-operative
  behavior without explicit inter-agent communication, and they
  avoided behavior which was individually desirable, but detrimental
  to the group's overall performance.  Furthermore, shaping the reward
  signal by explicitly penalizing certain patterns of sub-optimal
  behavior was found to dramatically improve the convergence rate.
\end{abstract}

\section{Introduction}

Network routing is the problem of efficiently using communication
paths. The standard example is Internet packet routing, but a broader
view incorporates messaging in multi-processor computers, and the
design of transport networks (such as vehicle traffic, airline
scheduling and water distribution). 

Current approaches to routing model the network as a graph with nodes
representing routers and weighted edges representing links and their
associated costs.  Finding good approaches to routing is thus reduced
to analytically solving various optimization problems for weighted
graphs \cite{deoPang84}. Although elegant solutions to such problems
have been found
\cite{dijkstra56,bellman58,cormenLeisersonRivest90,tanenbaum96}, the
algorithms can become intractable, as the models approach realism.  In
one example, \emcite{ordaRomSidi93} consider the problem of graphs
where the edge weights can change probabilistically (as a Markov
process).  In this case, they conclude that the minimal expected delay
routing problem is NP-hard.

Even if the problem posed by the model is tractable, unrealistic
modeling may mean that the derived solution is sub-optimal.  Real
world networks exhibit a variety of features which are not captured in
the standard weighted-graph model.  Such complexities include links
having limited bandwidth or capacity, limited buffering at either or
both ends of channels, packet collisions, mobile hosts (i.e.  changing
link delays), deadlock concerns, unreliable links, non-uniform traffic
(e.g. Poisson distributed, or bursty), minimum Quality-of-Service
requirements, and prioritized traffic \cite{tanenbaum96}.

There is also the problem of co-ordination, whether between routers or
between packets from one router.  Synchronization problems can occur
where two routers sharing two links simultaneously notice that one of
the links is not congested and both attempt to use it, causing
congestion.  This problem is worsened when the two routers notice
the other link is now free and both choose to use it, causing
congestion again.  This situation repeats, with the two routers
oscillating between the two links --- a phenomenon known as flapping
\cite{tanenbaum96}.

In view of the potential difficulties with model-based approaches to
network routing, in this paper we pursue an alternative, model-free
approach in which the routing problem is treated as a multi-agent
reinforcement learning problem. Each router is viewed as a single
independent agent, and makes its routing decisions according to a
local parameterized stochastic policy. The negative of the trip time
of each packet is used as an instantaneous reward signal delivered to
all agents (routers) upon arrival of the packet at its destination.
Each agent uses a policy-gradient reinforcement learning algorithm to
adjust the parameters of its policy in the direction of the gradient
of the average reward, leading to convergence of all router parameters
to a local maximum of the average reward, or equivalently to a local
minimum of the average packet trip time. The policy-gradient algorithm
we use is a multi-agent variant of the \textsc{Olpomdp} algorithm
introduced in \emcite{baxterWeaverBartlett99}, and analyzed in
\cite{drl_cdc2k}.

The key feature of multi-agent \textsc{Olpomdp} that makes it
attractive for the network routing problem is that the only non-local
information each router needs is the instantaneous reward signal
distributed upon a packet's arrival at its destination. In particular,
routers do not need to know the network topology or any other
information about the network in order to climb the gradient of the
global average reward.

We describe experiments on several different network models in which
multiple distributed routers learned to successfully minimize average
packet trip-time. The networks were designed such that routers had to
avoid individually desirable behaviour (minimizing trip-time for the
single router's packets alone) in order to achieve co-operative
behaviour that maximized the global objective. We also show that
shaping the reward signal by explicitly penalizing certain patterns of
sub-optimal behavior can dramatically improve the convergence rate,
alleviating one of the major difficulties with the use of stochastic
gradient algorithms.

\subsection{Reinforcement Learning and Routing}

The idea of applying reinforcement learning to network routing is not
new.  \emcite{boyanLittman93} introduced Q-routing, an adaptation of
Q-learning to network routing.  In this application, the destination
node is given, and the value of a state (i.e. node) is the expected
time-to-arrival from that node to the destination node.  Value
estimates are communicated between nodes, i.e. state evaluations are
updated based on values of future states.

This approach is very similar to Distance Vector routing, and the two
methods have the same roots.  Distance Vector routing is also known as
Bellman-Ford routing \cite{bellman58}, and Q-learning has its roots in
dynamic programming, also due to \emcite{bellman57}.

Although Q-routing adapted well to rising link costs (i.e. queueing
times), it has a number of weaknesses.  First, it is a deterministic
algorithm, in that at a particular time, there is only one link chosen
for traffic from a node.  If this link has limited bandwidth, then
routing all traffic through it may not be an optimal choice.  Second,
Q-routing does not maintain exploratory behavior --- when a link is
continually valued highest, then other links are never tried.  Network
changes (e.g. new hosts, faster links) may not be discovered, since a
sub-optimal policy can be continually followed whilst the value of a
better policy is under-estimated.  Third, the ``Count-to-Infinity''
problem may occur \cite{tanenbaum96}, when node \node{A}'s value is
calculated from node \node{B}'s value, which is calculated from node
\node{A}'s value.  This indirect self-reference can lead to long
convergence times.

\emcite{stone00} later applied TPOT-RL to network routing.  His
algorithm is explicitly multi-agent, and efficiently incorporates
domain-specific features such as link utilization rates into the
agents' observation space.  However, TPOT-RL learns value functions,
like Q-routing, and may suffer from the first two weaknesses outlined
above.  It does avoid the ``Count-to-Infinity'' problem, by using a
TD(1) style update rather than Q-routing's simpler TD(0) approach.

\section{Network Routing as a Multi-Agent, Partially Observable 
Markov Decision Process}

Each router in a network may be viewed as an agent that receives as
input packets destined for some other node in the network. The agent
(router) must decide on which of the possible outgoing links to send
the packet. Assuming a stochastic agent, let $\mu^i_u(y)$ denote the
probability that router $i$ chooses link $u$ for a packet routed to
destination node $y$. The decision problem is only partially
observable, since each agent sees only a subset of the packets, and
not the state of the entire network. It is a Markov process, because
the next state of the network is dependent only upon the current state
and the decisions of all the routers. Finally, multiple nodes makes it
a multi-agent problem. Thus, network routing is a multi-agent,
partially observable, Markov decision process (\textsc{Pomdp}).

To complete the picture we need to define a performance measure for
the network.  A natural performance measure at the packet level is
trip time, or the time taken for a packet to get from source node to
destination node.  Longer trip times are less desirable, so at time
$t$ we set the reward signal to be the negative sum of the trip times
of all packets which arrived at time $t$:
\begin{equation}
\label{eq:rdef}
r_t := - \sum_{p} {\mathit{trip\_time\_of}}( \, p \, )
\end{equation}
where the sum is over all packets that arrive at their destination at
time $t$.\footnote{ This is unrealistic, in that information at a node
  (i.e. the trip times of all arriving packets) has to be
  instantaneously communicated to all other nodes.  Alternatively,
  each node $i$ can calculate its component of the reward signal
  $r^i_t := - \sum_{p}{\mathit{trip\_time\_of}}( \, p \, )$ where now
  the sum is only over packets destined for $i$.  This component can
  be regularly broadcast via the network, and each router's reward
  signal is merely the sum of all components received in that time
  step. This means that rewards are delayed, but this does not affect
  a policy's \emph{long-term} average reward, although it may affect
  the training algorithm in that the delay between an action and its
  effect on the reward signal is increased. In practice, we found that
  modeling the broadcast and receipt of reward components via the
  network does not significantly affect results \cite{tao00}.}  Note
that provided the probability of each routing decision, $\mu^i_u(y)$,
is non-zero, every packet will eventually arrive at its destination
with probability 1.

To make the routers trainable, we parameterize each with a set of
real-valued parameters $\theta^i_{yu}$, one for each possible destination /
outgoing link pair $(y,u)$. To keep the computations simple, we use a Gibbs
distribution so that a packet arriving at router $i$, destined for
node $y$, is sent on link $u$ with probability
\begin{equation}
\label{eq:gibbs}
\mu^i_u(y, \theta) := \frac{e^{\theta^i_{yu}}}{\sum_{u'}
  e^{\theta^i_{yu'}}}
\end{equation} 
where the sum is over all outgoing links at router $i$.

\subsection{Multi-Agent Policy-Gradient Using OLPOMDP}

Our aim is to find parameter settings for all the routers that
maximizes the expected long-term average reward: 
\begin{equation}
\label{eq:avg_reward}
\eta(\theta) := {\mathbb E} \left[ \lim_{T\rightarrow\infty}
  \frac1T\sum_{t=1}^T r_t \right]
\end{equation}
where $\theta$ is the concatentaion of the parameters from all
routers.  Provided that the system is ergodic (i.e. it converges to a
unique steady state, given $\theta$), the right-hand-side of
\eqref{eq:avg_reward} is independent of the network starting state,
and converges with probability 1 over all possible reward sequences
$\{r_t\}$. Note that maximizing the long-term average reward
$\eta(\theta)$ is equivalent to minimizing the average packet
trip-time.

We use stochastic gradient ascent to find local optima of
$\eta(\theta)$. That is, at each time step $t$ the parameters
$\theta_t$ of all the routers are updated by 
\begin{equation}
\label{eq:theta_update}
\theta_{t+1} = \theta_t + \Delta\theta_t
\end{equation} 
where the long-term average of the updates $\Delta\theta_t$ lie in the
gradient direction $\nabla\eta(\theta)$. Since we are adjusting the
policy parameters $\theta$ to climb the gradient of the average reward
$\eta(\theta)$, this approach is known as a {\em policy-gradient}
algorithm. 

Policy-gradient algorithms were introduced into Machine Learning by
\emcite{williams92} with his \textsc{Reinforce} algorithm (similar
algorithms were described earlier in the Monte-Carlo literature by
\emcite{glynn86,reiman86}).  A practical difficulty with
\textsc{Reinforce}, and a number of subsequent algorithms
\cite{marbach98,baird98,meuleau99,meuleau00} is that they estimate the
gradient from a specified recurrent state, which requires perfect,
rather than partial, observability (at least in that state).  In
addition, the variance of the gradient estimate is related to the time
to return to this recurrent state, which can be large, particularly in
multiple-agent environments. For example, with network routing, the
time between identical configurations of the network can grow
exponentially with the size of the network.

Removing the need to identify a particular recurrent state,
\emcite{baxterBartlett99} introduced \textsc{Gpomdp}, an algorithm for
generating a biased estimate of the gradient
$\nabla_\beta\eta(\theta)$.  \textsc{Gpomdp} takes one free parameter
$\beta\in[0,1)$ which controls both the bias and the variance of the
estimates produced by the algorithm. In particular, the bias of
\textsc{Gpomdp} is small provided $\tau_{\textsc{alg}} := 1/(1-\beta)$
is small compared to a certain mixing time $\tau$ of a Markov process
associated with the \textsc{pomdp}. The relevant mixing time is
essentially the time constant in the decay of the influence of an
action on the reward signal. $\tau$ is always shorter than the
recurrence time, and is often considerably so. The variance of the
estimate produced by \textsc{Gpomdp} after $T$ steps is proportional
to $\tau_{\textsc{alg}} / T$, illustrating the bias/variance trade-off
in the selection of $\beta$: large $\beta$ gives small bias but large
variance and vice-versa. See \cite{drl_colt00} for a detailed
exposition.

\textsc{Olpomdp} is an online adaptation of \textsc{Gpomdp} which
applies policy parameter updates as soon as they are calculated
\cite{baxterWeaverBartlett99}.  If $\theta^i_t$ is the parameter
vector of the $i$-th router at time $t$, the updated parameter vector 
vector $\theta^i_{t+1}$ is given by 
\vspace*{-2mm}
\begin{equation}
\label{eq:theta_update1}
\theta^i_{t+1} := \theta^i_t + \gamma_t . r_t . z^i_t
\end{equation}
where $r_t$ is the sum of the rewards (negative trip times) for all
packets arriving at time $t$, $\gamma_t$ are suitable
step-sizes\footnote{Sufficient conditions on $\gamma_t$ are 
  $\gamma_t > 0$, $\sum\gamma_t = \infty$, $\sum\gamma_t^2 < \infty$,
  which is satisfied, for example,  by $\gamma_t = 1/t$, although to
  speed convergence in this paper we set $\gamma_t$ to be a suitably
  small constant.}, and $z^i_t$ is an eligibility trace of the same
dimensionality as $\theta_t^i$, which is updated according to 
\begin{equation}
\label{eq:z}
z^i_{t+1} := \beta . z^i_t +
  \frac{\nabla\mu^i_{u_t}(y_t, \theta^i)}{\mu^i_{u_t}(y_t, \theta^i)}
\end{equation}
Here $y_t$ is the destination node of the packet at router $i$ at time
$t$, and $u_t$ is the outgoing link on which it was routed.  The
gradient operator $\nabla$ is with respect to the parameters of the
$i$-th router, $\theta^i$.

With the Gibbs parameterization in \eqref{eq:gibbs}, 
\begin{equation}
\label{eq:nabla}
\frac{\nabla\mu^i_{u_t}(y_t, \theta^i)}{\mu^i_{u_t}(y_t,
  \theta^i)}
=
-[\,
{\scriptstyle
\mu^i_{1}(y_t, \theta^i),\, \dots,\,
\mu^i_{u_t}(y_t,\theta^i) -1,\, \dots,\,
\mu^i_{n}(y_t, \theta^i)
}\,]
\end{equation}

Since \textsc{Olpomdp} computes a biased estimate of the gradient, it
cannot be guaranteed to converge to a local optimum of the average
reward. However, provided the bias is sufficiently small (i.e $\beta$
is sufficiently large), \textsc{Olpomdp} will
converge to a region of near-zero gradient \cite{drl_cdc2k}. This can
be extended to the multi-agent case.

Apart from the globally distributed reward signal $r_t$, each router
uses only local information in its parameter updates (the router's
parameters, the packet destination, and the link on which the packet
is routed).  Thus, multi-agent \textsc{Olpomdp} is particularly
attractive as an algorithm for optimizing the co-operative behaviour
of all routers.  For similar observations in the context of episodic
tasks, see \cite{peshkin00}.

\section{Experiments}

\subsection{A Simple Network}
\label{result:simple}

\begin{figure}[ht]
  \vskip 0.2in
  \begin{center}
    \setlength{\epsfxsize}{1.00in}
    \centerline{\epsfbox{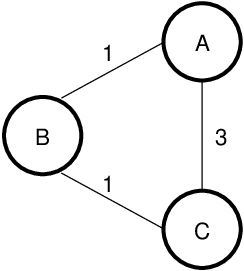}}
    \vskip 0.1in
    \caption{The triangle network.}
    \label{fig:3net}
  \end{center}
  \vskip -0.2in
\end{figure}

Figure \ref{fig:3net} depicts a simple network with three nodes, and
three links.  The numbers on the links are the link delay; the time
between when a packet is placed on the link and when it arrives at the
other end.  Note that the shortest path from \node{A} to \node{C} is
actually via \node{B}.  In this simulation, traffic was generated at a
rate of one packet per time step at each of the three nodes, with the
destination node chosen at random.

\begin{figure}[ht]
  \vskip 0.2in
  \begin{center}
    \setlength{\epsfxsize}{3.25in}
    \centerline{\epsfbox{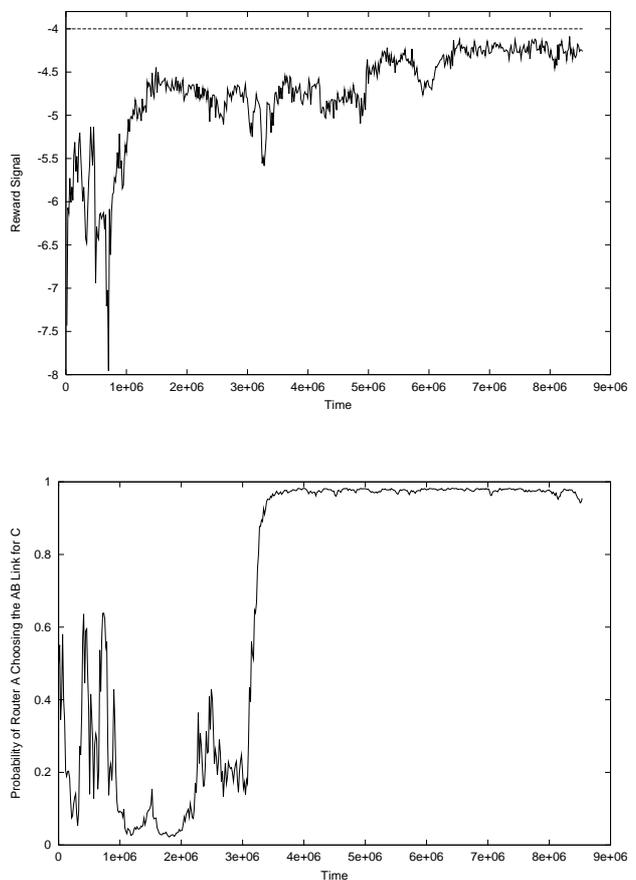}}
    \vskip 0.1in
    \caption{Reward signal $r_t$ and probability
      $\mu^{\mnode{A}}_{\mnode{AB}}({\mnode{C}})$ for
        \textsc{Olpomdp} on the triangle network.}
    \label{fig:3results}
  \end{center}
  \vskip -0.2in
\end{figure}

Figure \ref{fig:3results} shows a typical run of the \textsc{Olpomdp}
routing algorithm (with parameters $\beta = 0.99$ and $\gamma =
10^{-5}$) on the simple network.  The top chart shows that the reward
signal\footnote{The chart actually shows a moving average, since the
  reward signal is quite volatile.  Rewards are generated only when
  packets arrive at their destination, whilst traffic is generated
  probabilistically, and routed probabilistically.}  improves, on
average, over time to an optimum.  The bottom chart shows the
probability of the agent at \node{A} routing packets destined for
\node{C} on link \node{AB}.  Initially, it learns to avoid that link,
since the \node{AC} link (at a cost of 3 for certain) is preferable to
going via \node{B} (which chooses the wrong link half the time).
However, as \node{B} learns to route packets destined for \node{C},
the \node{AB} link (followed by the \node{BC} link) becomes a better
option.  The family of agents have learned the best collective policy,
even though it requires the agent at \node{A} to rely on the agent at
\node{B}.

\subsection{Mixed Strategies}
\label{result:mixed}

\begin{figure}[ht]
  \vskip 0.2in
  \begin{center}
    \setlength{\epsfxsize}{1.00in}
    \centerline{\epsfbox{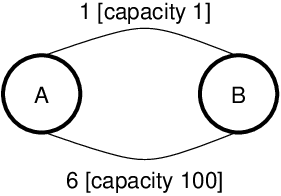}}
    \vskip 0.1in
    \caption{The contention network.}
    \label{fig:2net}
  \end{center}
  \vskip -0.2in
\end{figure}

Figure \ref{fig:2net} shows a two-node network, with two links from
\node{A} to \node{B}.  Unlike the previous example, these links have a
limited capacity, and packets placed on a saturated link are simply
dropped.  To capture the undesirability of losing packets, a penalty
term $d$ is subtracted from the reward signal every time a packet is
dropped.

In this experiment, traffic was only generated at \node{A} (and
destined for \node{B}), at a rate of two packets every time step.  If
the top link is chosen every time, then the average reward signal is
$-1-d$, since one packet gets through (for a trip time of 1) and one
packet is dropped.  If the bottom link is chosen every time, then the
average reward signal is $-12$, since two packets get through (for a
trip time of 6 each).  If a probabilistic policy is pursued, then the
long-term expected average reward can be calculated based on the
probabilities of each possible outcome occuring: both packets placed
on the top link, both on the bottom link, or one on each link.

If the drop penalty $d$ is set to 21, then a simple calculation shows
that the optimal probability of choosing the top link is
$\frac{1}{4}$.  Figure \ref{fig:2results} shows the performance of the
\textsc{Olpomdp} routing algorithm (with $\beta = 0.99$ and $\gamma =
10^{-7}$) on such a situation.  The routing agent has learned the
optimal non-deterministic policy, for a long-term average performance
of -10$\frac{3}{4}$.  This outperforms either deterministic strategy
(-22 for the top link only, and -12 for the bottom link only).

\begin{figure}[ht]
  \vskip 0.2in
  \begin{center}
    \setlength{\epsfxsize}{3.25in}
    \centerline{\epsfbox{figures/2NodeResults.eps}}
    \vskip 0.1in
    \caption{Reward signal $r_t$ and probability $\mu^{\mnode{A}}
      _{top\_link}({\mnode{B}})$ for
      \textsc{Olpomdp} on the 2-node contention network.}
    \label{fig:2results}
  \end{center}
  \vskip -0.2in
\end{figure}

\subsection{Penalizing Cycles}
\label{result:penalty}

The routing algorithm thus far does not include any domain knowledge.
In general, no shortest path can contain a cycle\footnote{This is
  because each extra link in a cycle taken incurs a further cost, but
  traversing a cycle has no net effect.  However, this result breaks
  down if links chosen for one packet can affect another packet, e.g.
  with limited bandwidth.}.  This fact has been exploited by
\emcite{diCaroDorigo98}, who drop all links of a cycle out of their
eligibility trace equivalent, and \emcite{stone00}, who immediately
downgrades an appropriate Q-value by some proportion of the duration
of the cycle.

In this paper, we incorporate an explicit global penalty term for
cycles.  The penalty is delivered as a negative component to the
reward signal.  This an example of reward shaping, where the
reinforcement signal $r$ consists of both an underlying performance
measure $r^{\mathit{underlying}}$ and a shaping term
$r^{\mathit{shaping}}$.  The two components are simply summed:
$$
r_t := r^{\mathit{underlying}}_t + r^{\mathit{shaping}}_t
$$
where the underlying performance measure is as before, given by
equation \ref{eq:rdef}.

Note that an optimal policy (under the original performance measure)
remains optimal even when cycles are penalized, since such a policy
cannot contain cycles

\begin{figure}[ht]
  \vskip 0.2in
  \begin{center}
    \setlength{\epsfxsize}{1.75in}
    \centerline{\epsfbox{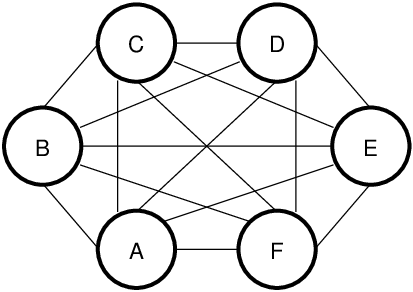}}
    \vskip 0.1in
    \caption{The complete six-node network.  All links have delay 1
    and unlimited capacity.}
    \label{fig:6net}
  \end{center}
  \vskip -0.2in
\end{figure}

Figure \ref{fig:6net} depicts a simple network with six nodes with
each link having delay 1 and unlimited capacity.  Packets maintained a
history of the last 2 nodes visited, and signalled a cycle if it
arrived at a node it had visited before.  The shaping term was set to
$-100$ multiplied by the number of cycles detected in that time step.

\begin{figure}[ht]
  \vskip 0.2in
  \begin{center}
    \setlength{\epsfxsize}{3.25in}
    \centerline{\epsfbox{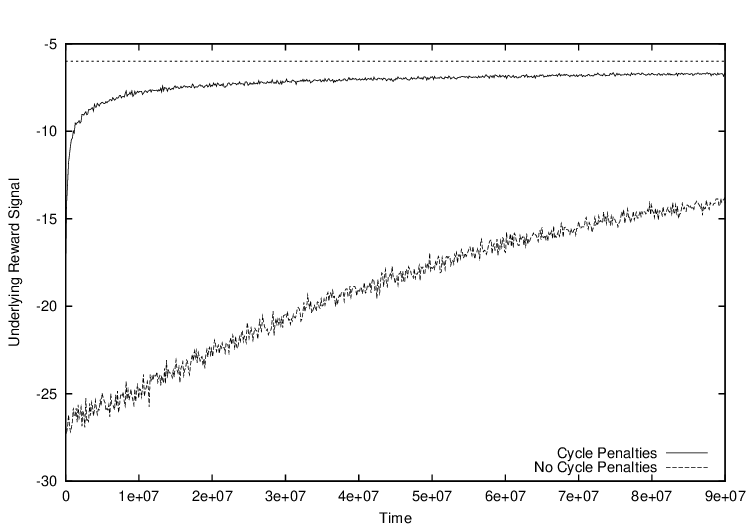}}
    \vskip 0.1in
    \caption{Improvement in the underlying reward signal, with
      and without cycle penalties, for \textsc{Olpomdp} on the
      complete six-node network.}
    \label{fig:6results}
  \end{center}
  \vskip -0.2in
\end{figure}

Figure \ref{fig:6results} shows the underlying performance measure for
two typical runs of the \textsc{Olpomdp} algorithm (with $\beta =
0.9$, and $\gamma = 10^{-6}$).  Clearly, the introduction of cycle
penalties has dramatically improved the convergence rate.  Note that
this occurred even without explicit credit assignment.  All agents
were penalized even if only one agent made the wrong decision.
Nonetheless, this scheme was very effective.

\subsection{Braess' Network Paradox}
\label{result:braess}

\emcite{braess68} discovered a routing problem in which adding an
extra path had the paradoxical effect of worsening overall
performance.  This paradox has been constructed for queueing networks
\cite{cohenKelly90}, but we take our formulation from
\emcite{tumerWolpert99}, which is simpler to analyze.

\begin{figure}[ht]
  \vskip 0.2in
  \begin{center}
    \setlength{\epsfxsize}{3.25in}
    \centerline{\epsfbox{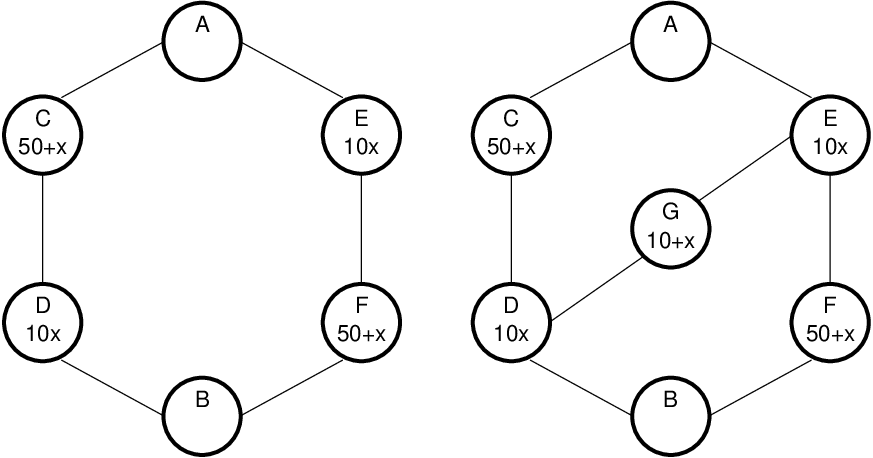}}
    \vskip 0.1in
    \caption{Braess' networks: the left hand network is $Braess_0$, and the
      right hand network is $Braess_1$.}
    \label{fig:braessnet}
  \end{center}
  \vskip -0.2in
\end{figure}

The two networks are shown in figure \ref{fig:braessnet}.  The network
model here differs from the previous experiments, in that costs are
incurred at nodes, rather than travelling through links.  The formulae
(e.g. $50 + x$) marking each node denote the cost per packet of using
that node, where $x$ is the traffic flow through that node.  The
motivation for this model comes from vehicular traffic, where each
additional packet (i.e. car) causes a cost (i.e. congestion and
waiting time) for each other packet through that node (i.e. cars at
that junction).

Consider the left hand network, which we denote $Braess_0$.  Suppose
that there are six packets to be routed from \node{A} to \node{B}.  If
all packets go down the left hand path, nodes \node{C} and \node{D}
record a flow of $x = 6$, and so the cost per packet is $(50 + 6) +
(10 \times 6) = 116$.  Similarly, if all packets are routed down the
right hand side, the total cost is $6 \times 116$.  In fact, the
lowest cost routing strategy is three packets down each path, for a
cost per packet of $(50 + 3) + (10 \times 3) = 83$.

Now consider the right hand network, denoted $Braess_1$.  Again, let
there be six packets to be routed from \node{A} to \node{B}.  Any
valid routing strategy for $Braess_0$ remains a valid routing strategy
for $Braess_1$, since $Braess_1$ is simply $Braess_0$ augmented by the
node \node{G} and the two links \node{EG} and \node{GD}.  In
particular, it is possible to achieve an average cost of $83$.

However, packets at \node{E} now have two paths to \node{B}: either
\node{EFB} or \node{EGDB}.  A greedy strategy (such as Q-routing) will
not stay at three packets down each path \node{ACDB} and \node{AEFB}
(and none down \node{AEGDB}).  This is because the cost from \node{E}
to \node{B} via \node{F} is an extra $53$ at \node{F}.  However, the
cost from \node{E} to \node{B} via \node{G} and \node{D} is $11$ at
\node{G} (with flow 1), and $40$ at \node{D} (with flow 4), which
totals to $51$.  Thus, the lowest cost path from \node{E} to \node{B}
for the marginal packet is actually through \node{G}.  In $Braess_1$,
equilibrium is achieved when two packets go down each path:
\node{ACDB}, \node{AEFB}, and \node{AEGDB}.  However, in this case,
the average cost per packet is $92$, which is higher than the best
strategy for $Braess_0$.

The \textsc{Olpomdp} routing algorithm (with $\beta = 0.99$ and
$\gamma = 10^{-5}$) was run on $Braess_1$, the augmented network.  In
this simulation, traffic was only generated at \node{A} for \node{B}
(at a rate of six packets per time step), and links were one-way and
downward (i.e.  \node{C} has only one link to choose from, and
\node{E} has two).  Figure \ref{fig:braessresults} shows a typical
run, with the top chart depicting the probability of packets at
\node{A} taking the left link, and the bottom chart depicting the
probability of packets at \node{E} taking the downward link.  Note
that the first number hovers around 50\%, whilst the second converges
to 100\%.  This represents the optimal policy, with half of the
traffic (i.e. three packets) going through \node{ACDB}, and of the
remaining three packets through \node{E}, zero go through \node{G},
with three going through \node{AEFB}.

\begin{figure}[ht]
  \vskip 0.2in
  \begin{center}
    \setlength{\epsfxsize}{3.25in}
    \centerline{\epsfbox{figures/braessResults.eps}}
    \vskip 0.1in
    \caption{Probabilities $\mu^{\mnode{A}}
      _{\mnode{AC}}({\mnode{B}})$ and
      $\mu^{\mnode{E}}
      _{\mnode{EF}}({\mnode{B}})$ for
      \textsc{Olpomdp} in Braess' second network.}
    \label{fig:braessresults}
  \end{center}
  \vskip -0.2in
\end{figure}

\section{Conclusion}

We have shown that \textsc{Olpomdp}, a policy-gradient reinforcement
learning algorithm, can successfully tackle a distributed, delayed
reward, partially observable, multiple agent learning problem.
Optimum solutions were found, and this was achieved without an
explicit system model (i.e. a set of system states and the state
transition function).

The algorithm was robust, performing well under a number of network
models: the link-delay-as-cost model (with and without link
capacities) of \S 3.1-3, and the node-flow-determines-cost model of \S
3.4.  This is in contrast to most routing algorithms that are
tailored to a specific network model and may fail if their assumptions
are broken.

In particular, we have shown that \textsc{Olpomdp}, and
policy-gradient methods in general, can
\begin{itemize}
  
\item learn co-operative behavior without explicit credit assignment,
  or even inter-agent communication other than the reward signal.
  (See \S \ref{result:simple})
  
\item learn mixed (or non-deterministic) strategies, when they are
  preferable to any deterministic strategy.  (See \S
  \ref{result:mixed})
  
\item learn at a much faster rate, when the reward signal is altered
  to explicitly penalize certain patterns of sub-optimal behavior.
  (See \S \ref{result:penalty})

\item learn to avoid behavior which is individually desirable, but
  detrimental to the group's performance.  (See \S
  \ref{result:braess})

\end{itemize}

These strengths are not limited to the domain of network routing.
Policy-gradient reinforcement learning is an effective approach with
wide applicability.

\section*{Acknowledgements} 
 
Nigel Tao received financial support from the Australian National
University and Burgmann College. This research was supported by the
Australian Research Council.

\bibliography{paper}
\bibliographystyle{mlapa}

\end{document}